\documentclass{sig-alternate-2013}

\newfont{\mycrnotice}{ptmr8t at 7pt}
\newfont{\myconfname}{ptmri8t at 7pt}
\let\crnotice\mycrnotice%
\let\confname\myconfname%

\permission{Permission to make digital or hard copies of all or part of this work for personal or classroom use is granted without fee provided that copies are not made or distributed for profit or commercial advantage and that copies bear this notice and the full citation on the first page. Copyrights for components of this work owned by others than the author(s) must be honored. Abstracting with credit is permitted. To copy otherwise, or republish, to post on servers or to redistribute to lists, requires prior specific permission and/or a fee. Request permissions from permissions@acm.org.}
\conferenceinfo{KDD'14,}{August 24--27, 2014, New York, NY, USA. \\
{\mycrnotice{Copyright is held by the owner/author(s). Publication rights licensed to ACM.}}}
\copyrightetc{ACM \the\acmcopyr}
\crdata{978-1-4503-2956-9/14/08\ ...\$15.00.\\
http://dx.doi.org/10.1145/2623330.2623635 }

\usepackage{subfigure} 
\usepackage{hyperref}
\usepackage{algorithm}
\usepackage{algorithmic}
\usepackage{fancyvrb}
\usepackage{verbatim}
\usepackage{wrapfig}
\usepackage{color}
\usepackage{stfloats}

\newcommand{\fullname}{Gradient Boosted Feature Selection}
\newcommand{\name}{GBFS}

\newcommand{\cut}[1]{}

\newcommand{\argmin}{\operatornamewithlimits{argmin}}
\newcommand{\x}{\mathbf{x}}

\newcommand{\w}{\mathbf{w}}

\newcommand{\wb}{\mathbf{w}}
\newcommand{\xb}{\mathbf{x}}

\newcommand{\h}{\mathbf{h}}

\newcommand{\bl}{\boldsymbol{\beta}}

\newcommand{\Fb}{\mathbf{F}}

\begin{document}

\title{Gradient Boosted Feature Selection}
% \subtitle{[Extended Abstract]
% \titlenote{A full version of this paper is available as
% \textit{Author's Guide to Preparing ACM SIG Proceedings Using
% \LaTeX$2_\epsilon$\ and BibTeX} at
% \texttt{www.acm.org/eaddress.htm}}}

\numberofauthors{4} %  in this sample file, there are a *total*
\author{
% 1st. author
\alignauthor
Zhixiang (Eddie) Xu \footnotemark[1]\\
       \affaddr{Washington University in St. Louis}\\
       \affaddr{One Brookings Dr.}\\
       \affaddr{St. Louis, USA}\\
       \email{xuzx@cse.wustl.edu}
% 2nd. author
\alignauthor
Gao Huang \\
       \affaddr{Tsinghua University}\\
       \affaddr{30 Shuangqing Rd.}\\
       \affaddr{Beijing, China}\\
       \email{huang-g09@mails.tsinghua.edu.cn}
% 3rd. author
\alignauthor Kilian Q. Weinberger\\
       \affaddr{Washington University in St. Louis}\\
       \affaddr{One Brookings Dr.}\\
       \affaddr{St. Louis, USA}\\
       \email{kilian@wustl.edu}
\and  % use '\and' if you need 'another row' of author names
% 4th. author
\alignauthor Alice X.\ Zheng \titlenote{Work done while at Microsoft Research}\\
       \affaddr{GraphLab}\\
       \affaddr{936 N. 34th St. Ste 208}\\
       \affaddr{Seattle, USA}\\
       \email{alicez@graphlab.com}
}
% There's nothing stopping you putting the seventh, eighth, etc.
% author on the opening page (as the 'third row') but we ask,
% for aesthetic reasons that you place these 'additional authors'
% in the \additional authors block, viz.
% \additionalauthors{Additional authors: John Smith (The Th{\o}rv{\"a}ld Group,
% email: {\texttt{jsmith@affiliation.org}}) and Julius P.~Kumquat
% (The Kumquat Consortium, email: {\texttt{jpkumquat@consortium.net}}).}
\date{13 June 2014}
% Just remember to make sure that the TOTAL number of authors
% is the number that will appear on the first page PLUS the
% number that will appear in the \additionalauthors section.

\maketitle
%!TEX root=gbfs.tex
\begin{abstract}
A feature selection algorithm should ideally satisfy four conditions: reliably extract relevant features; be able to identify non-linear feature interactions; scale linearly with the number of features and dimensions; allow the incorporation of known sparsity structure. In this work we propose a novel feature selection algorithm, \fullname{} (\name{}), which satisfies all four of these requirements. The algorithm is flexible, scalable, and surprisingly straight-forward to implement as it is based on a modification of Gradient Boosted Trees. We evaluate \name{} on several real world data sets and show that it matches or outperforms other state of the art feature selection algorithms.  Yet it scales to larger data set sizes and naturally allows for domain-specific side information. 
\end{abstract}

% Due to fast development in data collection and storage technologies, data sets are getting larger and larger. Feature selection in this large data set scenario becomes increasingly harder. In this paper, we introduce Greedy Nonlinear Feature Selection (GNFS), a novel and simple algorithm that employs gradient boosted regression trees to iteratively select subset of responsive features. As the algorithm builds on gradient boosted regression trees, not only does it provide nonlinear feature selection, but also naturally scales to very large data sets.
%  % We also propose to compute the variance of predictions given different number of features, which provides users most comprehensive information about the selected features. 
% We demonstrate the algorithm on several real-world data sets and show promising results.

%A category including the fourth, optional field follows...



\vfill\eject
\section{Introduction}
%!TEX root=gbfs.tex
Feature selection (FS)~\cite{guyon2003introduction} is an important problems in machine learning. In many applications, \emph{e.g.}, bio-informatics~\cite{saeys2007review} ~\cite{hellrung2012second} ~\cite{wang2015high} or neuroscience~\cite{liu2010decoding}, researchers hope to gain insight by analyzing \emph{how} a classifier can predict a label and what features it uses. Moreover, effective feature selection leads to parsimonious classifiers that require less memory~\cite{sra2011fast} and are faster to train and test~\cite{duchi2008efficient}.  It can also reduce feature extraction costs~\cite{ICML2013_xu13,xu2013anytime} and lead to better generalization~\cite{trevor2009elements}.

Linear feature selection algorithms such as LARS~\cite{friedman2001greedy} are highly effective at discovering linear dependencies between features and labels. However, they fail when features interact in nonlinear ways. Nonlinear feature selection algorithms, such as Random Forest~\cite{trevor2009elements} or recently introduced kernel methods~\cite{yamada2012high,song2012feature}, can cope with nonlinear interactions.  But their computational and memory complexity typically grow super-linearly with the training set size. As data sets grow in size, this is increasingly problematic. Balancing the twin goals of \emph{scalability} and \emph{nonlinear} feature selection is still an open problem.

In this paper, we focus on the scenario where data sets contain a large number of samples.  Specifically, we aim to perform efficient feature selection when the number of data points is much larger than the number of features ($n\!\gg\!d$). We start with the (NP-Hard) feature selection problem that also motivated LARS~\cite{friedman2001greedy} and LASSO~\cite{tibshirani1996regression}. But instead of using a linear classifier and approximating the feature selection cost with an $l_1$-norm, we follow~\cite{greedymiser} and use gradient boosted regression trees~\cite{friedman2001greedy} for which greedy approximations exist~\cite{breiman1984classification}. 

The resulting algorithm is surprisingly simple yet very effective. We refer to it as \fullname{} (\name{}). 
Following the gradient boosting framework, trees are built with the greedy CART algorithm~\cite{breiman1984classification}. Features are selected sparsely following an important change in the impurity function: splitting on new features is penalized by a cost $\lambda\!>\!0$, whereas re-use of previously selected features incurs no additional penalty.

\name{} has several compelling properties.
1. As it learns an ensemble of regression trees, it can naturally discover nonlinear interactions between features.
2. In contrast to, \emph{e.g.}, FS with Random Forests, it unifies feature selection and classification into a single optimization.
3. In contrast to existing nonlinear FS algorithms, its time and memory complexity scales as $O(dn)$, where $d$ denotes the number of features  dimensionality and $n$ the number of data points\footnote{In fact, if the storage of the input data is not counted, the memory complexity of \name{} scales as $O(n)$.}, and is very fast in practice.  
4. \name{} can naturally incorporate pre-specified feature cost structures or side-information, \emph{e.g.}, select bags of features or focus on regions of interest, similar to generalized lasso in linear FS~\cite{roth2004generalized}.

We evaluate this algorithm on several real-world data sets of varying difficulty and size, and we demonstrate that \name{} tends to match or outperform the accuracy and feature selection trade-off
of Random Forest Feature Selection, the current state-of-the-art in nonlinear feature selection. 

We showcase the ability of \name{} to naturally incorporate side-information  about inter-feature dependencies on a real world biological classification task~\cite{alon1999broad}. Here, features are grouped into {nine} pre-specified bags with biological meaning. \name{} can easily adapt to this setting and select entire feature bags. The resulting classifier matches the best accuracy of competing methods (trained on many features) with only \emph{a single} bag of features.

\section{Related Work}
%!TEX root=varfeat.tex
One of the most widely used feature selection algorithms is Lasso~\cite{tibshirani1996regression}. It minimizes the squared loss with $l_1$ regularization on the coefficient vector, which encourages sparse solutions.  Although scalable to very large data sets, Lasso models only linear correlations between features and labels and cannot discover non-linear feature dependencies. 

\cite{peng2005feature} propose the Minimum Redundancy Maximum Relevance (mRMR) algorithm, which selects a subset of the most responsive features that have high mutual information with labels. Their objective function also penalizes selecting redundant features. Though elegant, computing mutual information when the number of instance is large is intractable, and thus the algorithm does not scale.
HSIC Lasso~\cite{yamada2012high}, on the other hand, introduces non-linearity by combining multiple kernel functions that each uses a single feature. 
The resulting convex optimization problem aligns this kernel with a ``perfect'' label kernel. The algorithm requires constructing kernel matrices for all features, thus its time and memory complexity scale quadratically with input data set size. Moreover, both algorithms separate feature selection and classification, and require additional time and computation for training classifiers using the selected features.

Several other works avoid expensive kernel computation while maintaining non-linearity.  Grafting~\cite{perkins2003grafting} combines $l_1$ and $l_0$ regularization with a non-linear classifier based on a non-convex variant of the multi-layer perceptron. Feature Selection for Ranking using Boosted Trees~\cite{pan2009feature} selects the top features with the highest relative importance scores. \cite{tuv2009feature} and \cite{trevor2009elements} use Random Forest.
%by ranking all features from the forest by their accumulated impurity improvement in all splits. \cite{tuv2009feature} introduces artificial features and masking scores to Random Forest feature selection to further remove irrelevant features. 
Finally, while not a feature selection method, \cite{greedymiser} employ Gradient Boosted Trees to learn cascades of classifiers to reduce test-time cost by incorporating feature extraction budgets into the classifier optimization.

%However, different from our approach, these approaches select features by identifying feature importances from all weak learners in ensemble, whereas we find features by minimizing an optimization problem. 

% keerthi2005generalized
% This paper applies the generalized LARS algorithm to the SVM formulation with both L2 and L1 regularizer. Empirical study shows that this algorithm is efficient and to effective feature selection on text data sets.
% 
% kolter2009regularization
% This paper proposes a L1 regularization framework for the Least-Squares Temporal Difference (LSTD) algorithm which can be used for feature selection with high dimensional input.

% \section{Notation and Setup}
% \input{notation.tex}

%!TEX root=gbfs.tex
\section{Background}
Throughout this paper we type vectors in bold ($\x_i$), scalars in
regular math type ($k$ or $C$), sets in
cursive ($\mathcal{S}$) and matrices in capital bold ($\mathbf{F}$) font. Specific entries in vectors or matrices are
scalars and follow the corresponding convention. 

The data set consists of input vectors $\{\x_1,\dots,\x_n\}\in{\cal R}^d$ with corresponding labels $\{y_1,\dots,y_n\}\in {\cal Y}$ drawn from an unknown distribution. The labels can be binary,  categorical (multi-class) or real-valued (regression). For the sake of clarity, we focus on binary classification ${\cal Y}\in \{-1,+1\}$, although the algorithm can be extended to multi-class and regression as well.

% \begin{figure}[t!!!]
% \centerline{
% \includegraphics[width = .45\textwidth]{cappedl1}
% }
% \caption{The $l_1$ norm (cyan) and the capped $l_1$ norm with $\epsilon\!=\!1$ (red). The capped $l_1$ norm does not penalize the use of a feature beyond its initial extraction.  }
% \label{fig:cappedl1}
% \end{figure}

\subsection{Feature selection with the $l_1$ norm}

Lasso~\cite{tibshirani1996regression} combines linear classification and $l_1$ regularization
\begin{equation}
	\min_{\w} \sum_{(\x_i,y_i)} \ell(\x_i,y_i,\w) + \lambda |\w|_1. 
\end{equation}
In its original formulation, $\ell(\cdot)$ is defined to be the squared loss, $\ell(\x_i,y_i,\w)=(\wb^\top\x_i-y_i)^2$. However, for the sake of feature selection, other loss functions are possible.  In the binary classification setting, where $y_i\in\{-1,+1\}$, we use the better suited log-loss, 
%$\ell(\x_i,y_i,\w)=y_i\log(\wb^\top \x_i)+(1-y_i)\log(1-\wb^\top x_i)$. 
$\ell(\x_i,y_i,\w)=\log(1+\exp(y_i\wb^\top x_i))$~\cite{lee2006efficient}. 

\subsection{The capped $l_1$ norm}

$l_1$ regularization serves two purposes: It regularizes the classifier against overfitting, and it induces sparsity for feature selection. Unfortunately, these two effects of the $l_1$-norm are inherently tied and there is no way to regulate the impact of either one. 

\cite{CAPPEDLASSO08} introduce the \emph{capped} $l_1$ norm, defined by the element-wise operation
\begin{equation}
    q_\epsilon(w_i)=\min(|w_i|,\epsilon). \label{eqn:q_epsilon}
\end{equation}
%,  illustrated in figure~\ref{fig:cappedl1}. 
Its advantage over the standard $l_1$ norm is that once a feature is extracted, its use is not penalized further --- \emph{i.e.}, it penalizes using many features does not reward small weights. This is a much better approximation of the $l_0$ norm, which only penalizes feature use without interfering with the magnitude of the weights. 
When $\epsilon$ is small enough, \emph{i.e.}, $\epsilon\leq \min_i |w_i|$, we can compute the exact number of features extracted with $q_\epsilon(\wb)/\epsilon$. In other words, penalizing $q_\epsilon(\wb)$ is a close proxy for penalizing the number of extracted features. However, the capped $l_1$ norm is not convex and therefore not easy to optimize. 

The capped $l_1$ norm can be combined with a regular $l_1$ (or $l_2$) norm, where one can control the trade-off between feature extraction and regularization by adjusting the corresponding regularization parameters, $\mu, \lambda\geq 0$:
\begin{equation}
	\min_{\w} \sum_{(\x_i,y_i)} \ell(\x_i,y_i,\w) + \lambda |\w|_1 +  \mu q_\epsilon(\w).\label{eq:cappedlasso} 
\end{equation}
Here $q_\epsilon(\w)$ denotes $[q_\epsilon(\w_1), \ldots, q_\epsilon(\w_d)]$.

%!TEX root=gbfs.tex
\section{Gradient Boosted Feature Selection}

The classifier in Eq.~(\ref{eq:cappedlasso}) is better suited for feature selection than plain $l_1$ regularization.  However, it is still \emph{linear}, which limits the flexibility of the classifer. Standard approaches for incorporating non-linearity include the kernel learning~\cite{scholkopf2001learning} and boosting~\cite{chapelle2011boosted}. HSIC Lasso~\cite{yamada2012high} uses kernel learning to discover non-linear feature interactions at a price of quadratic memory and time complexity. 
Our method uses boosting, which is much more scalable.

Boosting assumes that one can pre-process the data with limited-depth regression trees. Let ${\cal H}$ be the set of all possible regression trees. Taking into account limited precision and counting trees that obtain identical values on the entire training set as one and the same tree, one can assume $|{\cal H}|$ to be finite (albeit possibly large). Assuming that inputs are mapped into ${\cal R}^{|{\cal H}|}$ through \mbox{$\phi(\xb)=[h_1(\xb),\dots,h_{|{\cal H}|}(\xb)]^\top$}, we propose to learn a linear classifier in this transformed space. Eq.~(\ref{eq:cappedlasso}) becomes
\begin{equation}
	\min_{\bl} \sum_{(\phi(\x_i),y_i)} \ell(\phi(\x_i),y_i,\bl) + \lambda |\bl|_1 +  \mu q_\epsilon(\bl).\label{eq:cappedlasso_bt}
\end{equation}
Here, $\bl$ is a sparse linear vector that selects trees.
Although it is extremely high dimensional, the optimization in Eq.~(\ref{eq:cappedlasso_bt}) is tractable because $\bl$ is extremely sparse. Assuming, without loss of generalization, that the trees in ${\cal H}$ are sorted so that the first $T$ entries of $\bl$ are non-zero, we obtain a final classifier
\begin{equation}
	H(\xb)=\sum_{t=1}^T \beta_t h_t(\xb). \label{eq:defH}
\end{equation}

\paragraph{Feature selection}
Eq.~(\ref{eq:cappedlasso_bt}) has two penalty terms: plain $l_1$ norm and capped $l_1$ norm. The first penalty term reduces overfitting while the second selects features. However, in its current form, the capped $l_1$ norm selects \emph{trees} rather than features. We therefore have to modify our setup to explicitly penalize the extraction of features.

To model the total number of features extracted by an ensemble of trees, we define a binary matrix $\Fb\in \{0,1\}^{d\times T}$, where an entry $F_{ft}=1$ if and only if the tree $h_t$ uses feature $f$. With this notation, we can express the total weight assigned to trees that extract feature $f$ as
\begin{align}
\sum_{t=1}^T |F_{ft}\beta_t|. \label{eq:totalfeat}
\end{align}
We modify $q_\epsilon(\bl)$ to instead penalize the actual weight assigned to \emph{features}.  The final optimization becomes
\begin{align}
	\min_{\bl} \ell(\bl) + \lambda |\bl|_1+ \mu\sum_{f=1}^d q_\epsilon\Bigg(\sum_{t=1}^T |F_{ft} \beta_t|\Bigg). \label{eq:optimization}
\end{align}
As before, if $\epsilon$ is sufficiently small (\mbox{$\epsilon\leq \min_f |\sum_{t=1}^T F_{ft}\beta_t|$}), we can set $\mu=1/\epsilon$ and the  feature selection penalty coincides exactly with the number of features used. 

%!TEX root=gbfs.tex
\subsection{Optimization}

The optimization problem in Eq.~(\ref{eq:optimization}) is non-convex and non-differentiable.  Nevertheless, we can minimize it effectively (up to a local fixed point) with gradient boosting~\cite{friedman2001greedy}. 
Let ${\cal L}(\bl)$ denote the loss function to be minimized and $\nabla {\cal L}(\beta)_t$ the gradient w.r.t $\beta_t$. Gradient boosting can be viewed as coordinate descent where we update the dimension with the steepest gradient at every step. We can assume that the set of all regression trees ${\cal H}$ is negation closed, \emph{i.e.}, for each $h\in{\cal H}$, we also have $-h\in{\cal H}$.  This  allows us to only follow negative gradients and always \emph{increase} $\bl$.  Thus there is always a non-negative optimal $\beta$. The search for the dimensions $t^*$ with the steepest negative gradient can be formalized as
\begin{align}
	t^* = \argmin_{t} \nabla{\cal L}(\beta)_t. \label{eq:optimalt}
\end{align}
In the remainder of this section we discuss approximate minimization strategies that does not require iterating over all possible trees. 

\paragraph{$l_1$-regularization}
Since each step of the optimization increases a single dimension of $\bl$ with a fixed step-size $\alpha>0$, the $l_1$ norm of $\bl$ can be written in closed form as $|\bl|_1=\alpha T$ after $T$ iterations. This means that penalizing the $l_1$ norm of $\bl$ is equivalent to early stopping after $T$ iterations~\cite{friedman2001greedy}. 
We therefore drop the $\lambda|\bl|_1$ term and instead introduce $T$ as an equivalent hyper-parameter. 

\paragraph{Gradient Decomposition}
To find the steepest descent direction at iteration $T'\!+\!1$, we decompose the (sub-)gradient into two parts, one for the loss function $\ell()$, and one for the capped $l_1$ norm penalty
\begin{align}
	\nabla {\cal L}(\bl)_t = \frac{\partial \ell}{\partial \beta_t} + \mu \sum_{f=1}^d\nabla q_\epsilon\Bigg(\sum_{t=1}^{T'} F_{ft}\beta_t \Bigg).
\end{align}
(Hereafter we drop the absolute value around $F_{ft}\beta_t$, since both $F_{ft}$ and $\beta_t$ are non-negative.)  The gradient of $q_\epsilon(\sum_t F_{ft}\beta_t)$ is not well-defined at the cusp when $\sum_t F_{ft}\beta_t=\epsilon$.  But we can take the right-hand limit, since $\beta_t$ never decreases,
\begin{align}
	\nabla q_\epsilon\Big(\sum_{t=1}^{T'} F_{ft}\beta_t\Big) = 
	\begin{cases}
     % \frac{1}{\epsilon}F_{ft}, & \textrm{if } |\sum_t F_{ft}\beta_t| < \epsilon \\
	 F_{ft}, & \textrm{if } \sum_t F_{ft}\beta_t < \epsilon \\
     0,	& \textrm{if } \sum_t F_{ft}\beta_t \ge \epsilon.
   \end{cases}	
\end{align}
If we set $\epsilon\!=\!\alpha$, where $\alpha\!>\!0$ is the step size, then $\sum_t F_{ft}\beta_t \ge \epsilon$ if and only if feature $f$ has already been used in a tree from a previous iteration. Let $\phi_f \!=\! 1$ indicate that feature $f$ is still \emph{unused}, and $\phi_f \!=\! 0$ otherwise. With this notation we can combine the gradients from the two cases and replace $\nabla q_\epsilon\Big(\sum_{t=1}^{T'} F_{ft}\beta_t\Big)$ with 
% $\frac{1}{\epsilon}\phi_f F_{ft}$. 
$\phi_f F_{ft}$. 
We obtain
\begin{align}
	% \nabla {\cal L}(\bl)_t = \frac{\partial \ell}{\partial \beta_t} + \frac{\mu}{\epsilon} \sum_{f=1}^d \phi_f F_{ft}.\label{eq:gradient2}
	\nabla {\cal L}(\bl)_t = \frac{\partial \ell}{\partial \beta_t} + \mu \sum_{f=1}^d \phi_f F_{ft}.\label{eq:gradient2}
\end{align}
Note that $\phi_fF_{ft}=1$ if and only if feature $f$ is extracted for the first time in tree $t$. In other words, the second term effectively penalizes trees that use many new (previously not selected) features. 

\paragraph{Greedy Tree construction} 
With Eq.~(\ref{eq:gradient2}) we can compute the gradient with respect to any tree.  But finding the optimal $t^*$ would still require searching all trees. In the remainder of this section, we transform the search for $t^*$ from a search over all possible dimensions $t$ to a search for the best tree $h_t$ to minimize a pre-specified loss function. The new search can be approximated with the CART algorithm~\cite{breiman1984classification}. 

To this end, we apply the chain rule and decompose $\frac{\partial \ell}{\partial \beta_t}$ into the derivative of the loss $\ell$ w.r.t.\ the current prediction evaluated at each input $H(\x_i)$ and the partial derivative $\frac{\partial H(\x_i)}{\partial \beta_t}$,
\begin{align}
	% \nabla {\cal L}(\beta)_t = \sum_{i=1}^n \frac{\partial \ell}{\partial H(\x_i)} \frac{\partial H(\x_i)}{\partial \beta_t} + \frac{\mu}{\epsilon} \sum_{f=1}^d \phi_f F_{ft}.
	\nabla {\cal L}(\beta)_t = \sum_{i=1}^n \frac{\partial \ell}{\partial H(\x_i)} \frac{\partial H(\x_i)}{\partial \beta_t} + \mu \sum_{f=1}^d \phi_f F_{ft}. \label{eq:chain}
\end{align}
Note that $H(\x_i) = \bl^\top \h(\x_i)$ is just a linear sum of all $h_t(\x_i)$, the predictions over training data.  Thus $\frac{\partial H(\x_i)}{\partial \beta_t} = h_t(\x_i)$. If we let $g_i$ denote the negative gradient $g_i = -\frac{\partial \ell}{\partial H(\x_i)}$, we can reformulate Eq.~(\ref{eq:chain}) as
\begin{align}
	% h_t = \argmin_{h_t \in {\cal H}} \sum_{i=1}^n -g_i h_t(\x_i) + \frac{\mu}{\epsilon} \sum_{f=1}^d \phi_f F_{ft}. \label{eq:gradientmatch}
	h_t = \argmin_{h_t \in {\cal H}} \sum_{i=1}^n -g_i h_t(\x_i) + \mu \sum_{f=1}^d \phi_f F_{ft}. \label{eq:gradientmatch}
\end{align}
Similar to~\cite{chapelle2011boosted}, we restrict ${\cal H}$ to only normalized trees (i.e. $\sum_i h_t^2(\x_i)=1$). We can then add two constant terms $\frac{1}{2}\sum_i h_t^2(\x_i)$ and $\frac{1}{2} \sum_i g_i^2$ to eq.~(\ref{eq:gradientmatch}), and complete the binomial equation.
\begin{align}
	% \hspace{-1ex}h_t = \argmin_{h_t \in {\cal H}} \frac{1}{2}\sum_{i=1}^n \Big(g_i - h_t(\x_i)\Big)^2 \!+\! \frac{\mu}{\epsilon} \sum_{f=1}^d \phi_f F_{ft}. \label{eq:impurity}
	\hspace{-1ex}h_t = \argmin_{h_t \in {\cal H}} \frac{1}{2}\sum_{i=1}^n \Big(g_i - h_t(\x_i)\Big)^2 \!+\! \mu \sum_{f=1}^d \phi_f F_{ft}.
	 \label{eq:impurity}
\end{align}
This is now a penalized squared loss---an \emph{impurity function}---and a good solutions can be found efficiently via the greedy CART algorithm for learning regression trees~\cite{friedman2001greedy}. The first term in Eq.~(\ref{eq:impurity}) encourages feature splits to best match the negative gradient of the loss function, and the second term rewards splitting on features which have already been used in previous iterations. 
Algorithm~\ref{table:algo} summarizes the overall algorithm in pseudo-code.
\begin{algorithm}[t!!!]
	\caption{\name{} in pseudo-code.\label{table:algo}}
	\label{algo}
	\begin{algorithmic}[1]
	\STATE Input: data $\{\x_i, y_i\}$, learning rate $\epsilon$, iterations $T$. \\
	\STATE Initialize predictions $H = 0$ and selected feature set $\Omega = \emptyset$.
	\FOR{$t = 1$ {\bf to} $T$}
	\STATE Learn $h_t$ using greedy CART to minimize the impurity function in Eq.~(\ref{eq:impurity}).
	\STATE Update $H = H + \epsilon h_t$.
	\STATE For each feature $f$ used in $h_t$, set $\phi_f = 0$ and $\Omega = \Omega \cup f$.
	\ENDFOR
	\STATE Return $H$ and $\Omega$.
	\end{algorithmic}
\end{algorithm}

%!TEX root=gbfs.tex
\subsection{Structured Feature Selection}
\label{sec:structure}
In many feature selection applications, one may have additional qualitative knowledge about acceptable sparsity patterns. Sometimes features can be grouped into bags and the goal is to select as few bags as possible. Prior work on handling structured sparsity include group lasso~\cite{huang2011learning,roth2008group} and generalized lasso~\cite{roth2004generalized}.  Our framework can easily handle structured sparsity via the feature cost identity function $\phi_f$. For example, we can define $\phi_f=1$ if and only if \emph{no} feature from the same bag as $f$ has been used in the past, and $\phi_f=0$ otherwise. The moment a feature from a particular bag is used in a tree, all other features in the same bag become ``free'' and the classifier is encouraged to use features from this bag exclusively until it starts to see diminishing returns. 

In the most general setting, we can define \mbox{$\phi_f:\Omega\rightarrow{\cal R}^+_0$} as a function that maps from the set of previously extracted features to a cost. For example, one could imagine settings where feature extraction appears in stages. Extracting feature $f$ makes feature $g$ cheaper, but not free.  One such application might be that of classifying medical images (e.g.,\ MRI scans) where the features are raw pixels and feature groups are local regions of interest. In this case, $\phi_f(\Omega)$ may reduce the ``price'' of pixels surrounding those in $\Omega$ to encourage feature selection with local focus.

%\input{method.tex}

% \section{Variance Analysis}
% \input{variance.tex}

\section{Results}
\label{sec:results}
%!TEX root=gbfs.tex

\begin{figure}[t!!!]
\centerline{
\includegraphics[width = .49\textwidth]{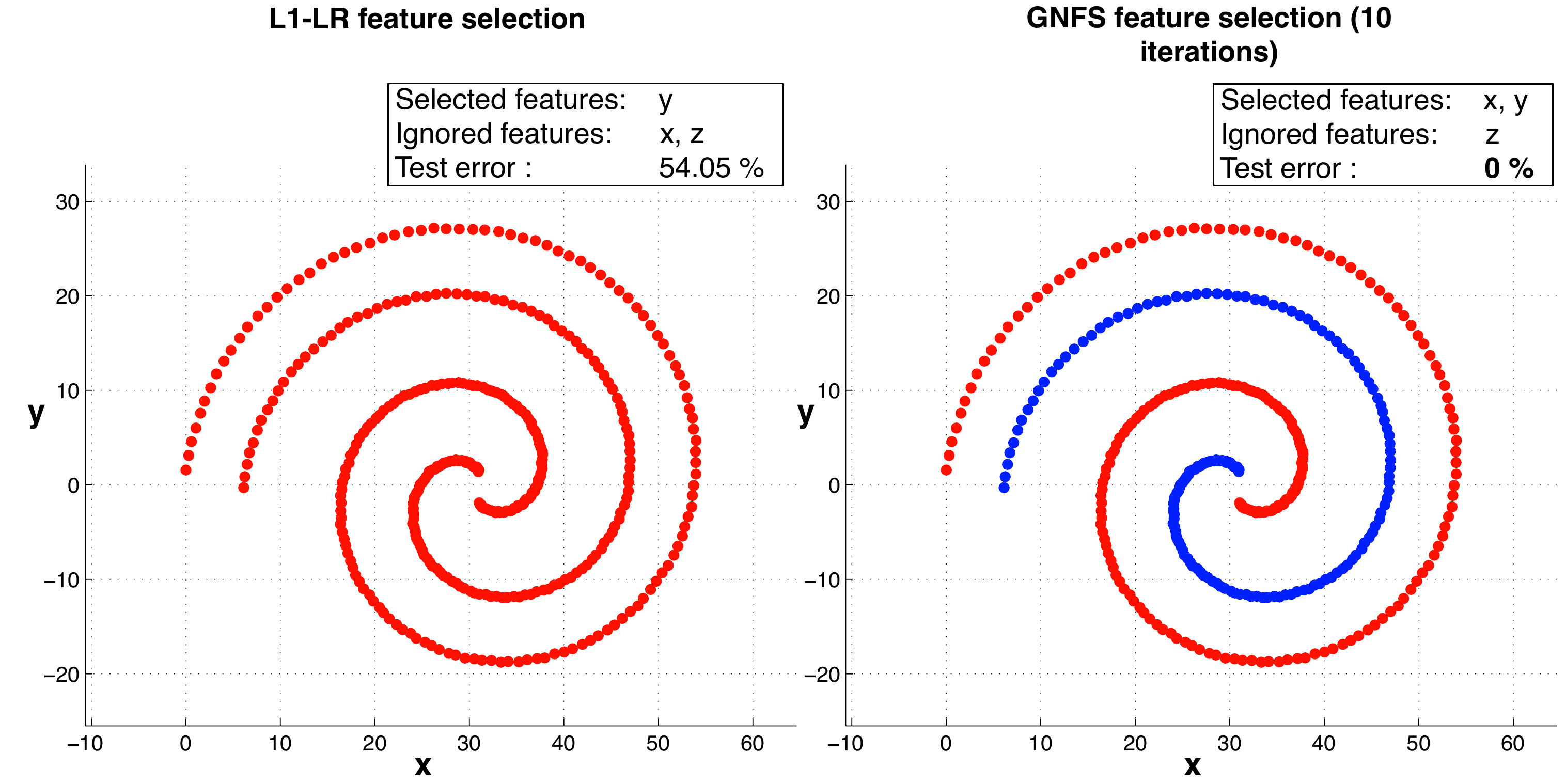}
}
\caption{Feature selection and classification performance on a simulated data set. \name{} clearly out-performs the $l_1$ regularized logistic regression as it successfully captures the nonlinear relations between labels and features.}
\label{fig:simul}
\end{figure}

In this section, we evaluate \name{} against other state-of-the-art feature selection methods on synthetic as well as real-world benchmark data sets.  We also examine at its capacity for dealing with known sparsity patterns in a bioinformatics application. All experiments were conducted on a desktop with dual 6-core Intel i7 cpus with 2.66GHz, 96 GB RAM, and Linux version 2.6.32.x86\_64.

\subsection{Synthetic data}
Figure~\ref{fig:simul} illustrates a synthetic binary classification data set with three features. The data is not linearly separable in either two dimensions or three dimensions. However, a good nonlinear classifier can easily separate the data using $x$ and $y$. The $z$ feature is simply a linear combination of $x$ and $y$ and thus redundant. We randomly select 90\% of the instances for training and the rest for testing. 
% Therefore, feature $x, y$ should be selected and $z$ should be detected as a redundant feature.

Figure~\ref{fig:simul} (left panel) illustrates results from \emph{$l_1$-regularized logistic regression} (L1-LR)~\cite{lee2006efficient,park2007l1}.  The regularization parameter is tuned on a hold-out set. 
Although L1-LR successfully detects and ignores the redundant feature $z$, it also assigns zero weight to $x$ and only selects a single feature $y$. Consequently, it has poor classification error rate on the test set ($54.05\%$). In contrast, \name{} (Figure~\ref{fig:simul}, right panel) not only identifies the redundant feature $z$, but also detects that the labels are related to a nonlinear combination of $x,y$. It selects both $x$ and $y$ and successfully separates the data, achieving $0\%$ classification error.

\begin{figure}[t!!!]
\centerline{
\includegraphics[width = .98\columnwidth]{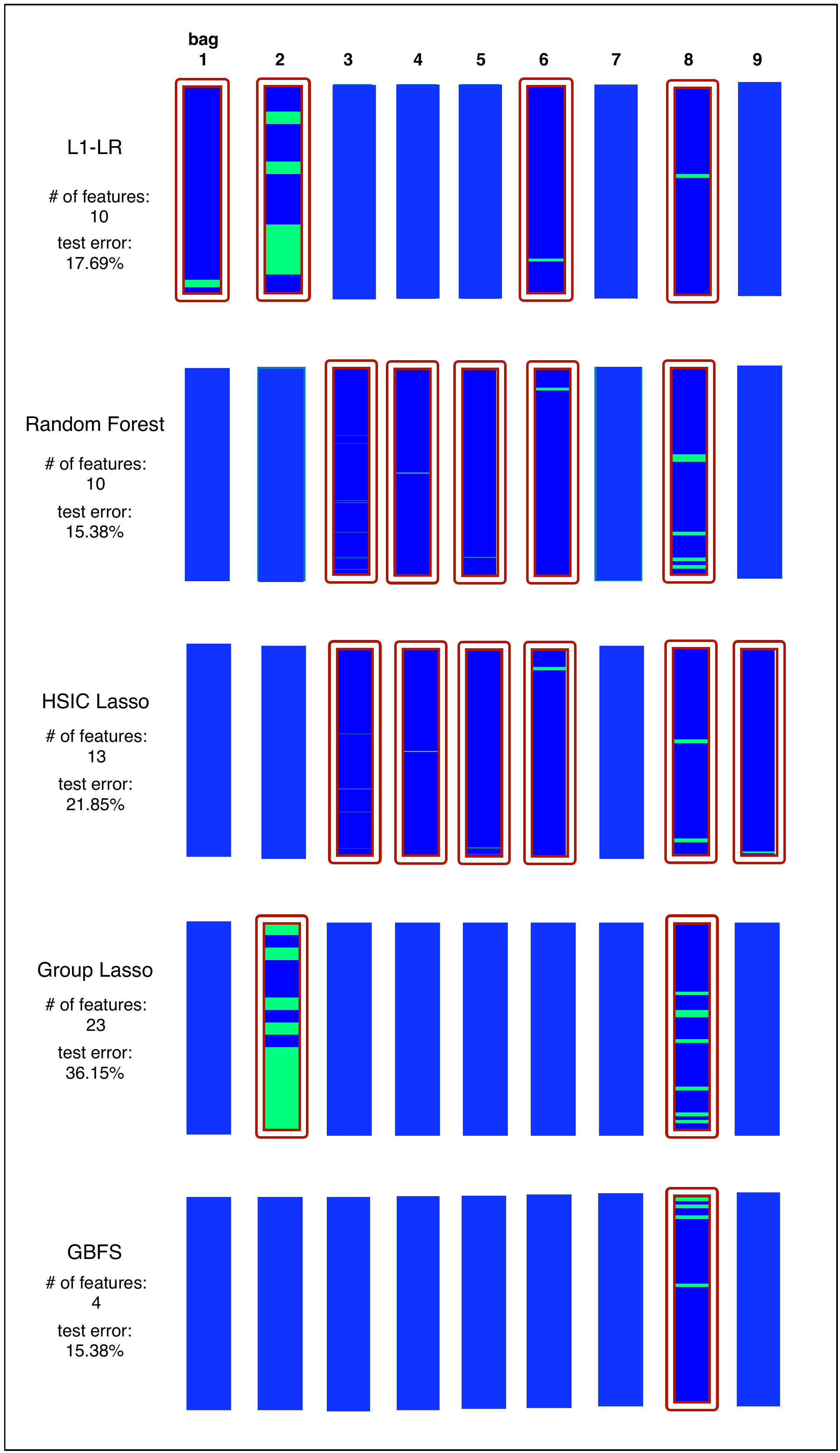}
}
% \vspace{-8pt}
\caption{Feature selection on structured feature data set. Selected features are colored in green, and unselected are in blue. The bag is highlighted with a red/white box if at least one of its features is selected by the classifier. (Some bags may require zooming in to make the selected features visible.)}
% \vspace{-10pt}
\label{fig:bag}
\end{figure}

\subsection{Structured feature selection}
In many applications there may be prior constraints on the sparsity patterns. 
Since \name{} can naturally incorporate pre-specified feature structures, we use it to perform structured feature selection on the Colon data set\footnote{Available through the Princeton University gene expression project (http://microarray.princeton.edu/oncology/)}. In this dataset, $40$ tumor and $22$ normal colon tissues for 6500 human genes are measured using affymetrix gene chips. \cite{alon1999broad} select 2000 genes that have the highest minimal intensity across the samples. \cite{ma2007supervised} further analyze these genes and cluster them into $9$ clusters/bags by their biological meaning. The task is to classify whether a tissue is normal or tumor. We random split the $62$ tissues into 80/20 training and testing datasets, repeated over $10$ random splits.
We use the feature-bag cost function $\phi_f$ mentioned in section~\ref{sec:structure} to incorporate this side-information (setting the cost of all features in a bag to \emph{zero} once the first feature is extracted). Feature selection without considering these bag information not only performs and 
generalizes poorly, but are also difficult to interpret and justify. 

Figure~\ref{fig:bag} shows the selected features from one random split and classification results averaged over $10$ splits. Selected features are colored in green, and unselected ones are in blue. A bag is highlighted with a red/white box if at least one of its features is selected by the classifier. We compare against $l_1$-regularized logistic regression (L1-LR)~\cite{lee2006efficient,park2007l1}, Random Forest feature selection (RF-FS)~\cite{trevor2009elements}, HSIC Lasso~\cite{yamada2012high} and Group Lasso~\cite{meier2008group}. 

As shown in Figure~\ref{fig:bag}, because \name{} can incorporate the bag structures, it focusses on selecting features in one specific bag. Throughout training, it only selects features from bag $8$ (highlighted with a red/white box). This conveniently reveals the association between diseases and gene clusters/bags. Similar to \name{}, Group Lasso with logistic regression can also deal with structured features. However, its $l_2$ regularization has side effects on feature weights, and thus results in much higher classification error rate $36.15\%$. In contrast, $l_1$-regularized logistic regression, Random Forest and HSIC Lasso do not take bag information into consideration.  They select scattered features from different bags, making results difficult to interpret. In terms of classification accuracy, \name{} and Random Forest has the lowest test set error rate ($15.38\%$), whereas $l_1$-regularized logistic regression (L1-LR) and HSIC Lasso achieve error rates of $17.69\%$ and $21.85\%$, respectively. 
%Our explanation why \name{}  other algorithms is two-fold: 1. it is indeed the case that the genes in bag $8$ are very predictive for testing if tissue is malignant or benign (a result that may be of high biological value); 2. \name{} does not penalize further feature extraction inside bag $8$ and can build a more accurate classifier than other methods, which keep penalizing all feature extractions.}

There are two reason why \name{} can be accurate with features from only a single bag.  First, it is indeed the case that the genes in bag $8$ are very predictive for the task of whether the tissue is malignant or benign (a result that may be of high biological value).  Second, \name{} does not penalize further feature extraction inside bag $8$ while other methods do; since bag 8 features are the most predictive, penalizing against them leads to a worse classifier.

% Moreover, the classification and feature selection results from \name{} is obtained only after $5$ iterations (trees), which is far from convergence. This demonstrates that \name{} can identify useful features even when it's not fully fit to the data. 

\begin{table*}[t]
	\begin{center}
\begin{tabular}{|l||c|c|c|c|c|c|c|}
  \hline
  \bf{data set} & pcmac & uspst & spam & isolet & mnist3vs8 & adult & kddcup99\\
  \hline \hline
  \bf{\#training} & 1555 & 1606 & 3681 & 6238 & 11982 & 32562 & 4898431 \\ 
  \hline
  \bf{\#testing} & 388 & 401 & 920 & 1559 & 1984 & 16282 & 311029\\
  \hline 
  \bf{\#features} & 3289 & 256 & 57 & 617 & 784 & 123 & 122 \\
  \hline
\end{tabular}
	\caption{Data sets statistics. Data sets are ordered by the number of training instances.}  \label{table:datasets}
	\end{center}
\end{table*}
\begin{figure*}[t!!!]
\centerline{
\includegraphics[width = 1\textwidth]{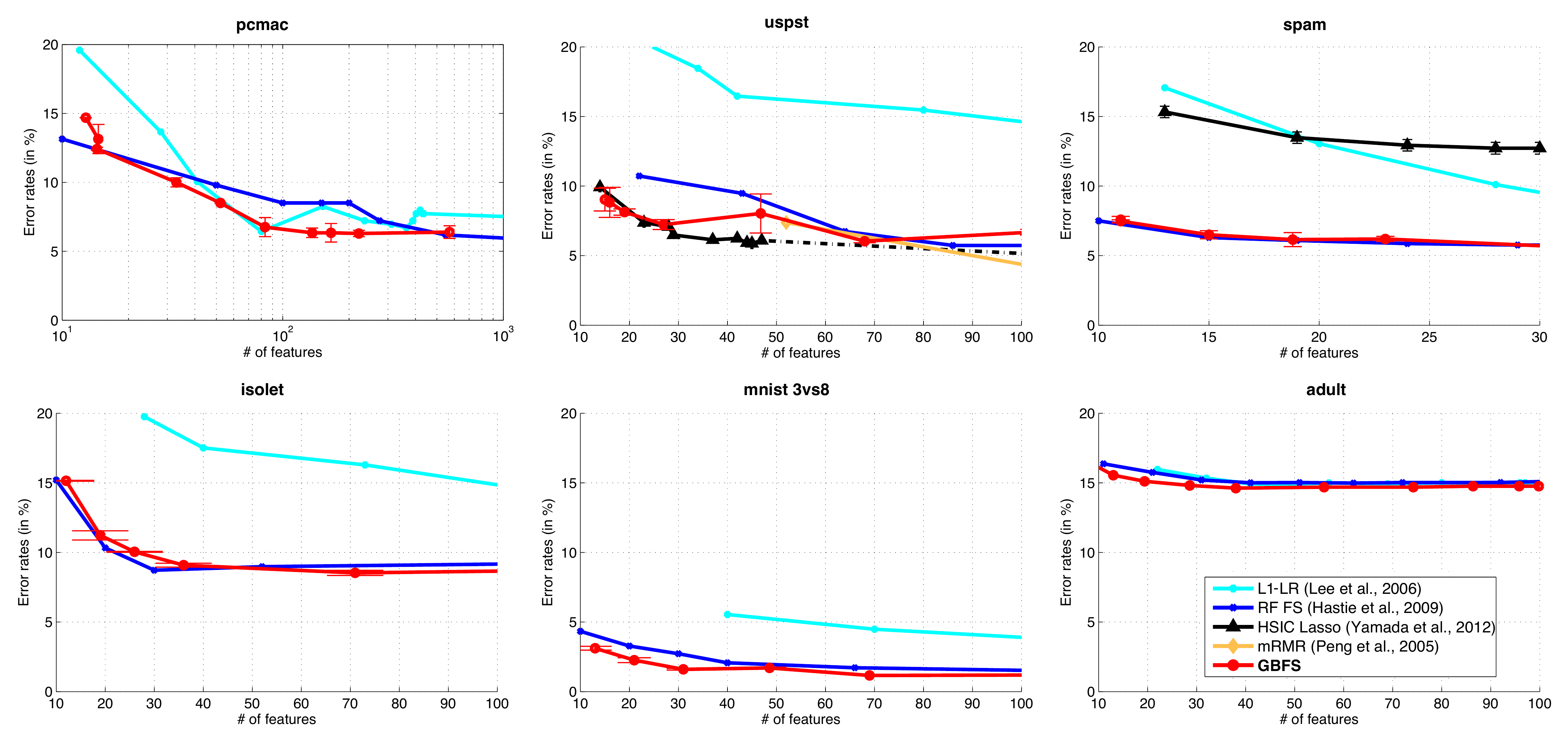}
}
\caption{Classification error rates (in \%) vs.\ feature selection performance for different algorithms on small to medium sized data sets.}
\label{fig:results}
\end{figure*}

\subsection{Benchmark data sets}
\paragraph{Data sets}
We evaluate \name{} on real-world benchmark data sets of varying sizes, domains and levels of difficulty. Table~\ref{table:datasets} lists data set statistics ordered by increasing numbers of training instances. We focus on data sets with a large number of training examples ($n\gg d$).
 % and leave the traditional feature selection data set ($d\gg n$)~\cite{zhao2010advancing} in the supplementary material.
All tasks are binary classification, though \name{} naturally extends to the regression setting, so long as the loss function is differentiable and continuous. Multi-class classification problems can be reduced to binary ones, either by selecting the two classes that are most easily confused or (if those are not known) by grouping labels into two sets.

\paragraph{Baselines} 
The first baseline is \emph{$l_1$-regularized logistic regression (L1-LR)}~\cite{lee2006efficient,park2007l1}.  We vary the regularization parameter to select different numbers of features and examine the test error rates under each setting.

We also compare against \emph{Random Forest feature selection (RF-FS)}~\cite{trevor2009elements}, a non-linear classification and feature selection algorithm. The learner builds many full decision trees by bagging training instances over random subsets of features. Features selection is done by ranking features based on their impurity improvement score, aggregated over all trees and all splits. Features with larger impurity improvements are considered more important. For each data set, we train a Random Forest with $2000$ trees and a maximum number of $20$ elements per leaf node. After training all $2000$ trees, we rank all features. Starting from top of the list, we re-train a random forest with only the top-k features and evaluate its accuracy on the test set. We increase the number of selected features until all features are included. 

The next state-of-the-art baseline is \emph{Minimum Redundancy Maximum Relevance (mRMR)}~\cite{peng2005feature}, a non-linear feature selection algorithm that ranks features based on their mutual information with the labels. Again, we increase the selected feature set starting from the top of the list.  At each stopping point, we train an RBF kernel SVM using only the features selected so far. The hyper-parameters are tuned on on 5 different random 80/20 splits of the training data.  The final reported test error rates are based on the SVM trained on the full training set with the best hyper-parameter setting. 
%Doe to scalability limitations, we were only able to run mRMR on the smallest data set (\emph{uspst}). 

Finally, we compare against HSIC Lasso~\cite{yamada2012high}, a convex extension to Greedy HSIC~\cite{song2012feature}. HSIC Lasso builds a kernel matrix for each feature and combines them to best match an ideal kernel  generated from the labels. It encourages feature sparsity via an $l_1$ penalty on the linear combination coefficients. Similar to $l_1$-regularized logistic regression, we evaluate a wide range of $l_1$ regularization parameters to sweep out the entire feature selection curve. Since HSIC Lasso is a two steps algorithm, we train a kernel SVM  with the selected features to perform classification. Similar to the mRMR experiment, we use cross-validation to select  hyper-parameters and average over $5$ runs. 
%HSIC scales super-linearly with respect to the data set size $n$ and number of features $d$, and therefore could only be evaluated on two data sets (\emph{uspst}, \emph{spam}). 

To evaluate \name{}, we perform 5 random 80/20 training/validation splits. We use the validation set to choose the depth of the regression trees and the number of iterations (maximum iterations is set to $2000$). The learning rate is set to $\epsilon=0.1$ for all data sets. In order to show the whole error rates curve, we evaluate the algorithm at $10$ values for the feature selection trade-off parameter $\mu$ in Eq.~(\ref{eq:optimization}) (i.e., $\mu = \{2^{-3},2^{-2},2^{-1},2^0,2^1,2^2,2^{3},2^{5},2^7,2^9\}$). 

\paragraph{Error rates}
Figure~\ref{fig:results} shows the feature selection and classification performance of different algorithms on small and medium sized data sets. 
We select up to 100 features except for \emph{spam} ($d=57$) and \emph{pcmac} ($d=3289$). 
In general, $l_1$-regularized logistic regression (L1-LR), Random Forest (RF-FS) and \name{} easily scale to all data sets. RF-FS and \name{} both clearly out perform L1-LR in accuracy on all data sets due to their ability to capture nonlinear feature-label relationships. HSIC Lasso is very sensitive to the data size (both the number of training instance and the number of features), and only scales to two small data sets (\emph{uspst,spam}). mRMR is even more restrictive (more sensitive to the number of training instance) and thus only works for \emph{uspst}. Both of them run out of memory on \emph{pcmac}, which has the largest number of features. In terms of accuracy, \name{} clearly outperforms HSIC Lasso on \emph{spam} but performs slightly worse on \emph{uspst}. 
%Accuracy wise, except for \emph{uspst}, \name{} out-performs HSIC Lasso. 
% Note that HSIC Lasso and mRMR achieving high classification accuracy results from applying RBF kernel SVM as the classifier. 
On all small and medium datasets, \name{} either out-performs RF-FS or matches its classification performance. However, very different from RF-FS, \name{} is a one step approach that selects features and learns a classifier at the same time, whereas RF-FS requires re-training a classifier after feature selection. This means that \name{} is much faster to train than RF-FS.

\begin{figure}[t!!!]
\centerline{
\includegraphics[width = 0.98\columnwidth]{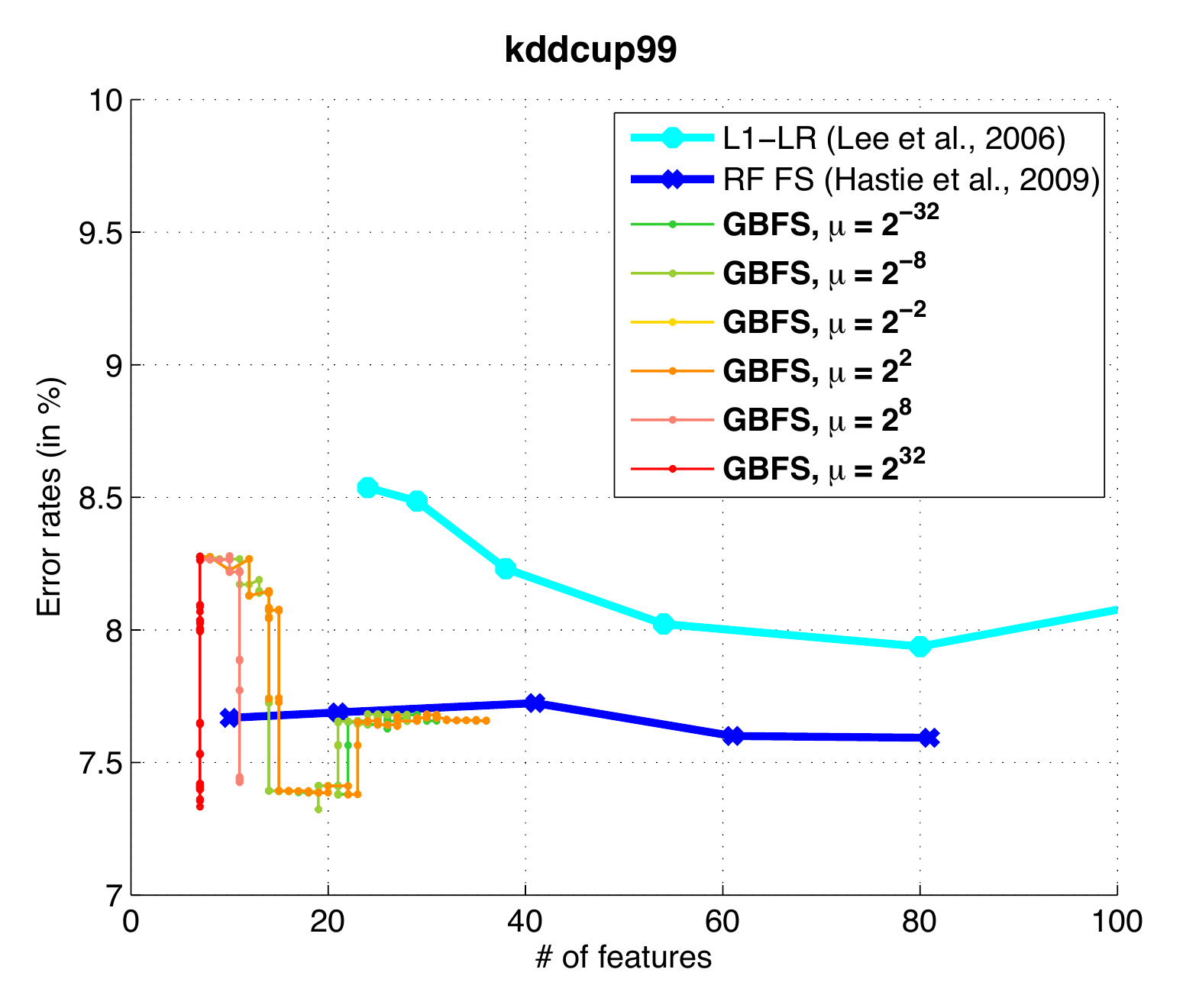}
}
\caption{Feature selection and classification error rates (in \%) for different algorithms on the large \emph{kddcup99} data set.}
% \vspace{-15pt}
\label{fig:largeset}
\end{figure}

\begin{figure}[t!!!]
\centerline{
\includegraphics[width = .49\textwidth]{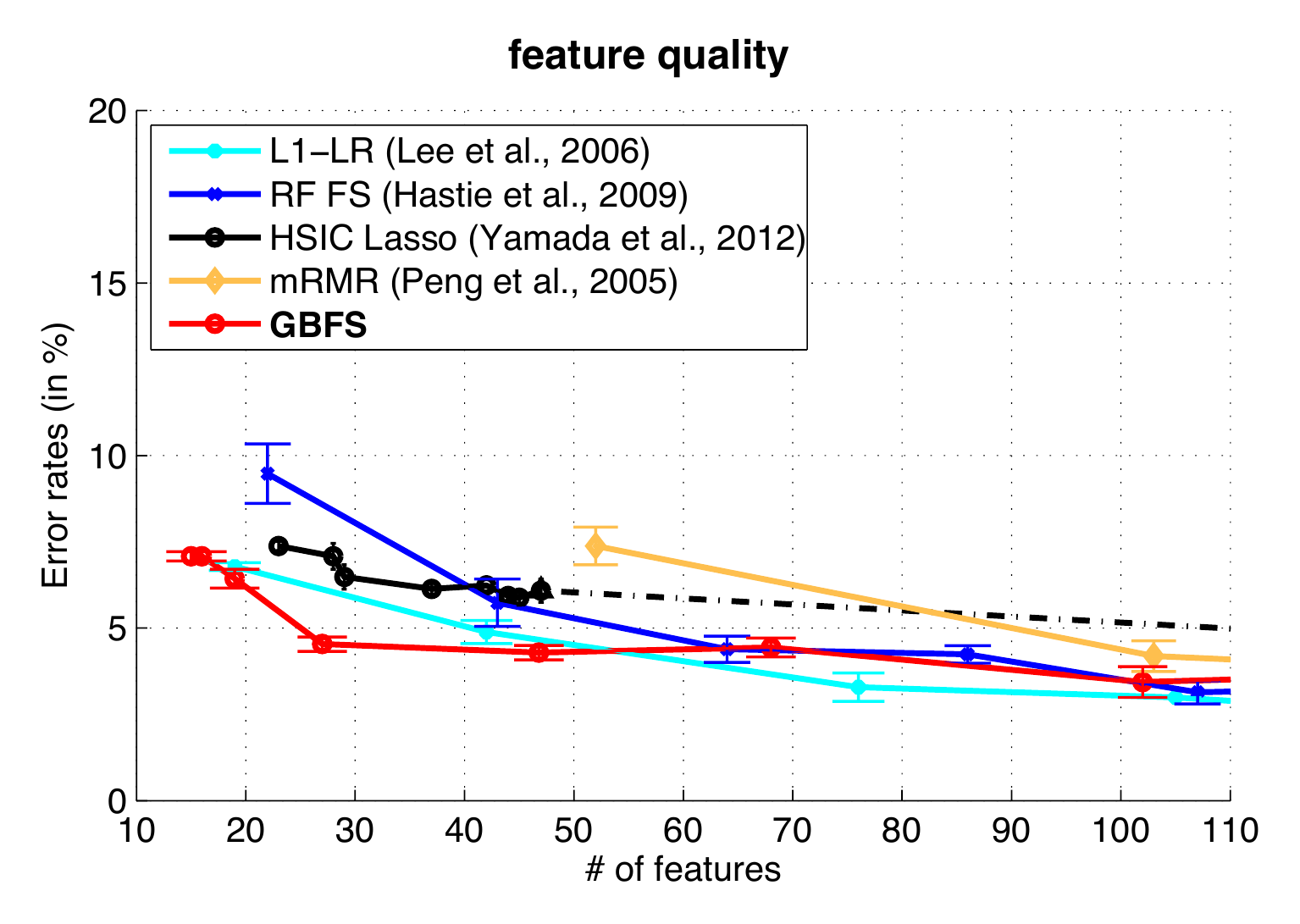}
}
\caption{Error rates (in \%) of SVM-RBF trained on various feature subset obtained with different features selection algorithms.}
% \vspace{-18pt}
\label{fig:quality}
\end{figure}

\paragraph{Large data set} 
The last dataset in Table~\ref{table:datasets} (\emph{kddcup99}) contains close to 5 million training instances. Training on such large data sets can be very time-consuming. We limit \name{} to $T=500$ trees with the default hyper-parameters of tree depth = 4 and learning rate = 0.1. Training Random Forest with default hyper-parameters would take more than a week. Therefore, we limit the number of trees to $100$ and the maximum number of instances per leaf node to $500$. Feature selection and classification results are shown in Figure~\ref{fig:largeset}. For \name{}, instead of plotting the best performing results for each $\mu$, we plot the whole feature selection iteration curve for multiple values of $\mu$. \name{} obtains lower error rates than Random Forest (RF-FS) and $l_1$ regularized logistic regression (L1-LR) when few features are selected. 
(Note that due to the extremely large data set size, even improvements of $<1\%$ are considered significant.)

\paragraph{Quality}
To evaluate the quality of the selected features, we separate the contribution of the feature selection from the effect of using different classifiers. 
We apply all algorithms on the smallest data set (\emph{uspst}) to select a subset of the features and then train a SVM with RBF kernel on the respective feature subset. Figure~\ref{fig:quality} shows the error rates as a function of the number of selected features. \name{} obtains the lowest error rates in the (most interesting) regions of only few selected features. As more features are selected eventually all FS algorithms converge to similar values. It is worth noting that the linear classifier (L1-LR) slightly outperforms most nonlinear methods when given enough features, which suggests that the uspst digits data requires a nonlinear classifier for prediction but not for feature discovery. 

\begin{figure}[t!!!]
\centerline{
\includegraphics[width = .49\textwidth]{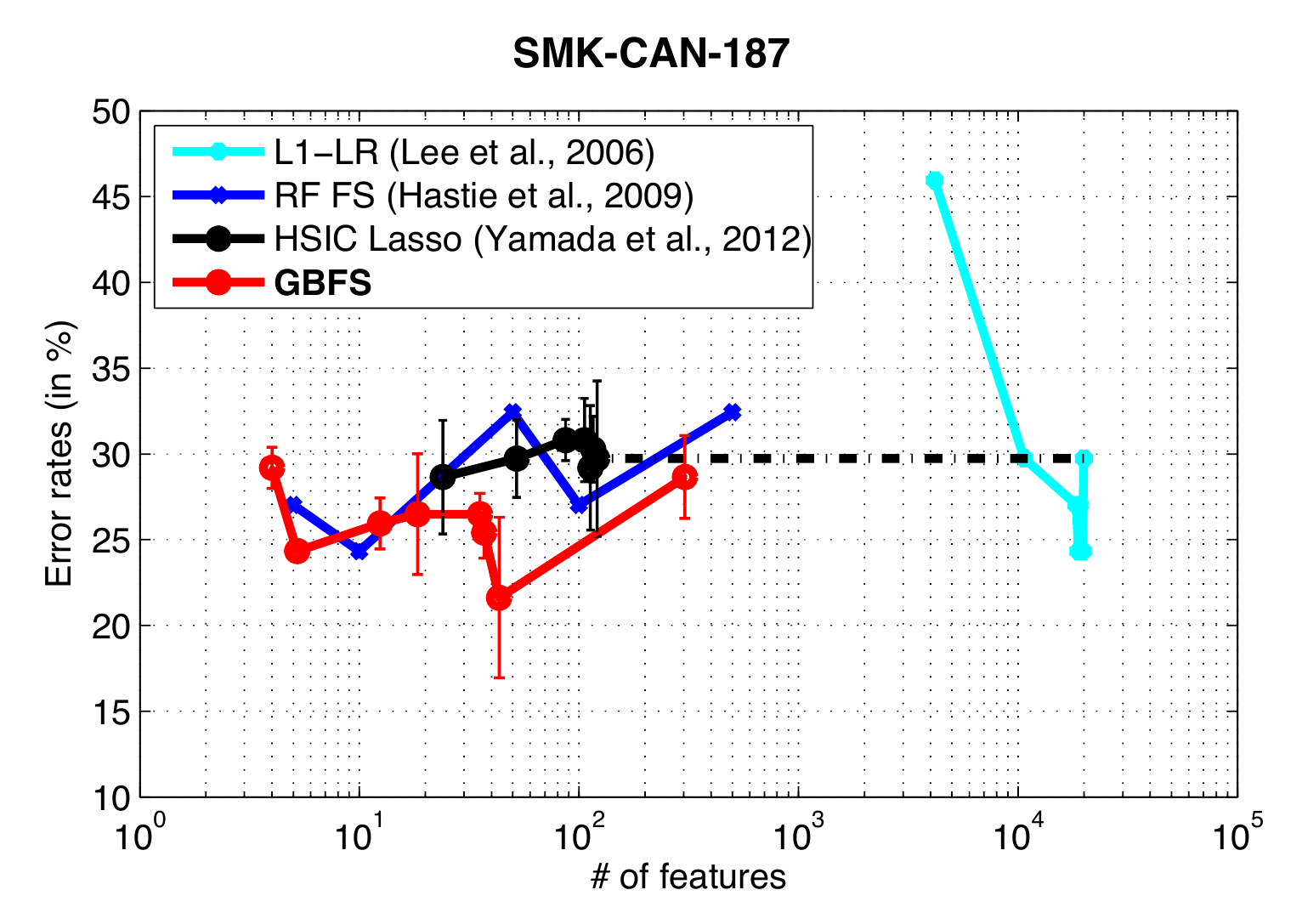}
}
\caption{Classification error rates (in \%) vs. feature selection performance for different algorithms on a high dimensional ($d\gg n$) data set.}
% \vspace{-18pt}
\label{fig:high_dims}
\end{figure}

\paragraph{$d\gg n$ scenario}
While \name{} focusses on the scenario where the number of training data points is much larger than the number of features ($n\gg d$), we also evaluate \name{} on a traditional feature selection benchmark data set SMK-CAN-187~\cite{spira2007airway}, which is publicly available from~\cite{zhao2010advancing}. This binary classification data set contains $187$ data points and $19,993$ features. We average results over 5 random $80/20$ train-test splits. Figure~\ref{fig:high_dims} compares the results. \name{} out-performs $l_1$-regularized logistic regression (L1-LR), HSIC-Lasso and Random Forest feature selection (RF-FS), though by a small margin in some regions.

% \begin{figure}[t!!!]
% \centerline{
% \includegraphics[width = 1.1\columnwidth]{timing}
% }
% \caption{Training run-time for different algorithms on small to medium sizes datasets.}
% \label{fig:timing}
% \end{figure}

\paragraph{Computation time and complexity}
Not surprisingly, the linear method (L1-LR) is the fastest by far.
Both mRMR and HSIC Lasso take significantly more time than Random Forest and \name{} because they involve either mutual information or kernel matrix computation, which scales as $O(d^2)$ or $O(n^2)$. Random Forest builds full trees, requiring a time complexity of $O(\sqrt{d}n\log n)$ per tree. The  dependency on $\sqrt{d}$ is slightly misleading, as the number of trees required for Random Forests is also dependent on the number of features and scales as $O(\sqrt{d})$ itself. 
In contrast, \name{} only builds limited depth (depth = $4,5$) trees, and the computation time complexity is  $O(dn)$. The number of iterations $T$ is independent of the number of input features $d$; it is only a function of how the number of desired features. 
Empirically, we observe that the two algorithms are comparable in speed but \name{} is significantly faster on data sets with many instances (large $n$). The training time ranged from several seconds to minutes on the small data sets to about one hour on the large data  set \emph{kddcup99} (when Random Forest is trained with only 500 trees and large leaf sizes). 
Admittedly, the empirical comparison of training time is slightly problematic because our Random Forest implementation is based on highly optimized C++ code, whereas \name{} is implemented in Matlab\texttrademark. We expect that \name{} could be made significantly faster if implemented in faster programming languages (\emph{e.g.} C++) with the incorporation of known parallelization techniques for limited depth trees~\cite{tyree2011parallel}. 
%Comparing training time between \name{} and Random Forests is difficult for two reasons: For Random Forest we used a highly optimized C++ optimization, whereas \name{} was implemented in Matlab\texttrademark. Further, both algorithms can be run for any number of trees, thus trading off time complexity for feature quality. Overall, Random Forests and \name{} require about the same amount of time

%Therefore, on small data sets (\emph{uspst, spam}), \name{} is $3$X faster than Random Forest, and on medium size data sets, \name{} is faster by a factor of $5$ to $10$. For the large data set \emph{kddcup99}, \name{} takes $2.2$ hours to train $500$ trees, and Random Forest takes $6.2$ hours to train $100$ trees with minimum $500$ instances in leaf nodes, which performs worse than \name{}.

%!TEX root=gbfs.tex
\section{Discussion}
This paper introduces \name{}, a novel algorithm for non-linear feature selection. 
The algorithm quickly train very accurate classifiers while selecting high quality features. 
In contrast to most prior work, \name{} is based on a variation of gradient boosting of  limited depth trees~\cite{friedman2001greedy}. This approach has several advantages. It scales naturally to large data sets and it combines learning a powerful classifier and performing feature selection into a single step. It can easily incorporate known feature dependencies, a common setting in biomedical applications~\cite{alon1999broad}~\cite{wang2013discontinuous}, medical imaging~\cite{etzel2009introduction}~\cite{wang2015high} and computer vision~\cite{dalal2005histograms}. This has the potential to unlock interesting new discoveries in a variety of application domains. 
%In particular, we are interested in exploring the use of \name{} in settings with more intricate sparsity structures. Examples may be classification on fMRI scans~\cite{etzel2009introduction} and other medical applications. 
From a practitioners perspective, it is now worth investigating if a data set has inter-feature dependencies that could be provided as additional side-information to the algorithm. 

One bottleneck of \name{} is that it explores features using the CART algorithm, which has a complexity of $O(dn)$. This may become a problem in cases with millions of features. Although this paper primarily focusses on the $n\gg d$ scenario, as future work we plan to consider improving the scalability with respect to $d$. One promising approach is to restrict the search to a random subsets of new features, akin to Random Forest. However, in contrast to Random Forest, the iterative nature of \name{} allows us to bias the sampling probability of a feature by its splitting value from previous iterations---thus avoiding unnecessary selection of unimportant features. 
 
We are excited by the promising results of \name{} and 
believe that the use of gradient boosted trees for feature selection will lead to many interesting follow-up results. This will hopefully spark new algorithmic developments and improved feature discovery across application domains. 

%This flexibility opens up interesting new research directions within data mining, feature selection and also application domains. 
%In conclusion, we introduced a novel feature selection algorithm, \name{}. 

\section{Acknowledgments}
KQW was supported by NSF grants 1149882 and 1137211. KQW and ZEX were supported by NSF IIS-1149882 and IIS- 1137211. Part of this work was done while ZEX was an intern at Microsoft Research, Redmond.

\bibliographystyle{abbrv}
\bibliography{gbfs}  % sigproc.bib is the name of the Bibliography in this case

\begin{thebibliography}{10}

\bibitem{alon1999broad}
U.~Alon, N.~Barkai, D.~A. Notterman, K.~Gish, S.~Ybarra, D.~Mack, and A.~J.
  Levine.
\newblock Broad patterns of gene expression revealed by clustering analysis of
  tumor and normal colon tissues probed by oligonucleotide arrays.
\newblock {\em Proceedings of the National Academy of Sciences},
  96(12):6745--6750, 1999.

\bibitem{breiman1984classification}
L.~Breiman.
\newblock {\em Classification and regression trees}.
\newblock Chapman \& Hall/CRC, 1984.

\bibitem{chapelle2011boosted}
O.~Chapelle, P.~Shivaswamy, S.~Vadrevu, K.~Weinberger, Y.~Zhang, and B.~Tseng.
\newblock Boosted multi-task learning.
\newblock {\em Machine learning}, 85(1):149--173, 2011.

\bibitem{dalal2005histograms}
N.~Dalal and B.~Triggs.
\newblock Histograms of oriented gradients for human detection.
\newblock In {\em Computer Vision and Pattern Recognition, 2005. CVPR 2005.
  IEEE Computer Society Conference on}, volume~1, pages 886--893. IEEE, 2005.

\bibitem{duchi2008efficient}
J.~Duchi, S.~Shalev-Shwartz, Y.~Singer, and T.~Chandra.
\newblock Efficient projections onto the $l1$-ball for learning in high
  dimensions.
\newblock In {\em Proceedings of the 25th international conference on Machine
  learning}, pages 272--279. ACM, 2008.

\bibitem{etzel2009introduction}
J.~A. Etzel, V.~Gazzola, and C.~Keysers.
\newblock {An introduction to anatomical ROI-based fMRI classification
  analysis}.
\newblock {\em Brain Research}, 1282:114--125, 2009.

\bibitem{friedman2001greedy}
J.~Friedman.
\newblock {Greedy function approximation: A gradient boosting machine}.
\newblock {\em The Annals of Statistics}, pages 1189--1232, 2001.

\bibitem{guyon2003introduction}
I.~Guyon and A.~Elisseeff.
\newblock An introduction to variable and feature selection.
\newblock {\em The Journal of Machine Learning Research}, 3:1157--1182, 2003.

\bibitem{trevor2009elements}
T.~Hastie, R.~Tibshirani, and J.~H. Friedman.
\newblock {\em The elements of statistical learning}.
\newblock Springer, 2009.

\bibitem{hellrung2012second}
J.~L. Hellrung~Jr, L.~Wang, E.~Sifakis, and J.~M. Teran.
\newblock A second order virtual node method for elliptic problems with
  interfaces and irregular domains in three dimensions.
\newblock {\em Journal of Computational Physics}, 231(4):2015--2048, 2012.

\bibitem{huang2011learning}
J.~Huang, T.~Zhang, and D.~Metaxas.
\newblock Learning with structured sparsity.
\newblock {\em The Journal of Machine Learning Research}, 12:3371--3412, 2011.

\bibitem{lee2006efficient}
S.~Lee, H.~Lee, P.~Abbeel, and A.~Y. Ng.
\newblock Efficient l1 regularized logistic regression.
\newblock In {\em Proceedings of the National Conference on Artificial
  Intelligence}, volume~21, page 401. Menlo Park, CA; Cambridge, MA; London;
  AAAI Press; MIT Press; 1999, 2006.

\bibitem{liu2010decoding}
Y.~Liu, M.~Sharma, C.~Gaona, J.~Breshears, J.~Roland, Z.~Freudenburg,
  E.~Leuthardt, and K.~Q. Weinberger.
\newblock Decoding ipsilateral finger movements from ecog signals in humans.
\newblock In {\em Advances in Neural Information Processing Systems}, pages
  1468--1476, 2010.

\bibitem{ma2007supervised}
S.~Ma, X.~Song, and J.~Huang.
\newblock Supervised group lasso with applications to microarray data analysis.
\newblock {\em BMC bioinformatics}, 8(1):60, 2007.

\bibitem{meier2008group}
L.~Meier, S.~Van De~Geer, and P.~B{\"u}hlmann.
\newblock The group lasso for logistic regression.
\newblock {\em Journal of the Royal Statistical Society: Series B (Statistical
  Methodology)}, 70(1):53--71, 2008.

\bibitem{pan2009feature}
F.~Pan, T.~Converse, D.~Ahn, F.~Salvetti, and G.~Donato.
\newblock Feature selection for ranking using boosted trees.
\newblock In {\em Proceedings of the 18th ACM conference on Information and
  knowledge management}, pages 2025--2028. ACM, 2009.

\bibitem{park2007l1}
M.~Y. Park and T.~Hastie.
\newblock L1-regularization path algorithm for generalized linear models.
\newblock {\em Journal of the Royal Statistical Society: Series B (Statistical
  Methodology)}, 69(4):659--677, 2007.

\bibitem{peng2005feature}
H.~Peng, F.~Long, and C.~Ding.
\newblock Feature selection based on mutual information criteria of
  max-dependency, max-relevance, and min-redundancy.
\newblock {\em Pattern Analysis and Machine Intelligence, IEEE Transactions
  on}, 27(8):1226--1238, 2005.

\bibitem{perkins2003grafting}
S.~Perkins, K.~Lacker, and J.~Theiler.
\newblock Grafting: Fast, incremental feature selection by gradient descent in
  function space.
\newblock {\em The Journal of Machine Learning Research}, 3:1333--1356, 2003.

\bibitem{roth2004generalized}
V.~Roth.
\newblock The generalized lasso.
\newblock {\em Neural Networks, IEEE Transactions on}, 15(1):16--28, 2004.

\bibitem{roth2008group}
V.~Roth and B.~Fischer.
\newblock {The group-lasso for generalized linear models: Uniqueness of
  solutions and efficient algorithms}.
\newblock In {\em Proceedings of the 25th international conference on Machine
  learning}, pages 848--855. ACM, 2008.

\bibitem{saeys2007review}
Y.~Saeys, I.~Inza, and P.~Larra{\~n}aga.
\newblock A review of feature selection techniques in bioinformatics.
\newblock {\em bioinformatics}, 23(19):2507--2517, 2007.

\bibitem{scholkopf2001learning}
B.~Sch{\"o}lkopf and A.~Smola.
\newblock {\em Learning with kernels: Support vector machines, regularization,
  optimization, and beyond}.
\newblock MIT press, 2001.

\bibitem{song2012feature}
L.~Song, A.~Smola, A.~Gretton, J.~Bedo, and K.~Borgwardt.
\newblock Feature selection via dependence maximization.
\newblock {\em The Journal of Machine Learning Research}, 98888:1393--1434,
  2012.

\bibitem{spira2007airway}
A.~Spira, J.~E. Beane, V.~Shah, K.~Steiling, G.~Liu, F.~Schembri, S.~Gilman,
  Y.-M. Dumas, P.~Calner, P.~Sebastiani, et~al.
\newblock Airway epithelial gene expression in the diagnostic evaluation of
  smokers with suspect lung cancer.
\newblock {\em Nature medicine}, 13(3):361--366, 2007.

\bibitem{sra2011fast}
S.~Sra.
\newblock Fast projections onto $l1$, $q$-norm balls for grouped feature
  selection.
\newblock In {\em Machine Learning and Knowledge Discovery in Databases}, pages
  305--317. Springer, 2011.

\bibitem{tibshirani1996regression}
R.~Tibshirani.
\newblock Regression shrinkage and selection via the lasso.
\newblock {\em Journal of the Royal Statistical Society. Series B
  (Methodological)}, pages 267--288, 1996.

\bibitem{tuv2009feature}
E.~Tuv, A.~Borisov, G.~Runger, and K.~Torkkola.
\newblock Feature selection with ensembles, artificial variables, and
  redundancy elimination.
\newblock {\em The Journal of Machine Learning Research}, 10:1341--1366, 2009.

\bibitem{tyree2011parallel}
S.~Tyree, K.~Weinberger, K.~Agrawal, and J.~Paykin.
\newblock Parallel boosted regression trees for web search ranking.
\newblock In {\em WWW}, pages 387--396. ACM, 2011.

\bibitem{wang2013discontinuous}
L.~Wang and P.-O. Persson.
\newblock A discontinuous galerkin method for the navier-stokes equations on
  deforming domains using unstructured moving space-time meshes.
\newblock In {\em 21st AIAA Computational Fluid Dynamics Conference}, page
  2833, 2013.

\bibitem{wang2015high}
L.~Wang and P.-O. Persson.
\newblock A high-order discontinuous galerkin method with unstructured
  space--time meshes for two-dimensional compressible flows on domains with
  large deformations.
\newblock {\em Computers \& Fluids}, 118:53--68, 2015.

\bibitem{ICML2013_xu13}
Z.~Xu, M.~K., M.~Chen, and K.~Q. Weinberger.
\newblock Cost-sensitive tree of classifiers.
\newblock In S.~Dasgupta and D.~Mcallester, editors, {\em Proceedings of the
  30th International Conference on Machine Learning (ICML-13)}, volume~28,
  pages 133--141. JMLR Workshop and Conference Proceedings, 2013.

\bibitem{xu2013anytime}
Z.~Xu, M.~Kusner, G.~Huang, and K.~Q. Weinberger.
\newblock Anytime representation learning.
\newblock In {\em Proceedings of the 30th International Conference on Machine
  Learning (ICML-13)}, pages 1076--1084, 2013.

\bibitem{greedymiser}
Z.~Xu, K.~Weinberger, and O.~Chapelle.
\newblock The greedy miser: Learning under test-time budgets.
\newblock In {\em ICML}, pages 1175--1182, 2012.

\bibitem{yamada2012high}
M.~Yamada, W.~Jitkrittum, L.~Sigal, E.~P. Xing, and M.~Sugiyama.
\newblock High-dimensional feature selection by feature-wise non-linear lasso.
\newblock {\em arXiv preprint arXiv:1202.0515}, 2012.

\bibitem{CAPPEDLASSO08}
T.~Zhang.
\newblock Multi-stage convex relaxation for learning with sparse
  regularization.
\newblock In D.~Koller, D.~Schuurmans, Y.~Bengio, and L.~Bottou, editors, {\em
  Advances in Neural Information Processing Systems 21}, pages 1929--1936.
  2008.

\bibitem{zhao2010advancing}
Z.~Zhao, F.~Morstatter, S.~Sharma, S.~Alelyani, A.~Anand, and H.~Liu.
\newblock Advancing feature selection research.
\newblock {\em ASU Feature Selection Repository}, 2010.

\end{thebibliography}
% ACM needs 'a single self-contained file'!
%\balancecolumns
\appendix
\section{Supplementary results}
%!TEX root=gbfs.tex
We further extend our experimental results by incorporating more one-vs-one pairs from MNIST data set. We randomly pick $6$ one-vs-one pairs from MNIST and evaluate \name{} and other feature selection algorithms. The first baseline is $l_1$-regularized logistic regression (L1-LR).  We vary the regularization parameter to select different number of features and examine the error rates under these different settings. We also compare against \emph{Random Forest feature selection}~\cite{trevor2009elements}. Same to the procedure described in section~\ref{sec:results}, we run Random Forest with $2000$ trees and a maximum number of $20$ elements per leaf node. After training all $2000$ trees, we rank all features. Starting from the most important feature, we re-train a random forest with only selected features and evaluate it on testing set. We gradually include less important features until we include all features. The other two baselines (include mRMR, HSIC-Lasso) do not scale on the MNIST data set.

\begin{figure*}[t!!!]
\centerline{
\includegraphics[width = .9\textwidth]{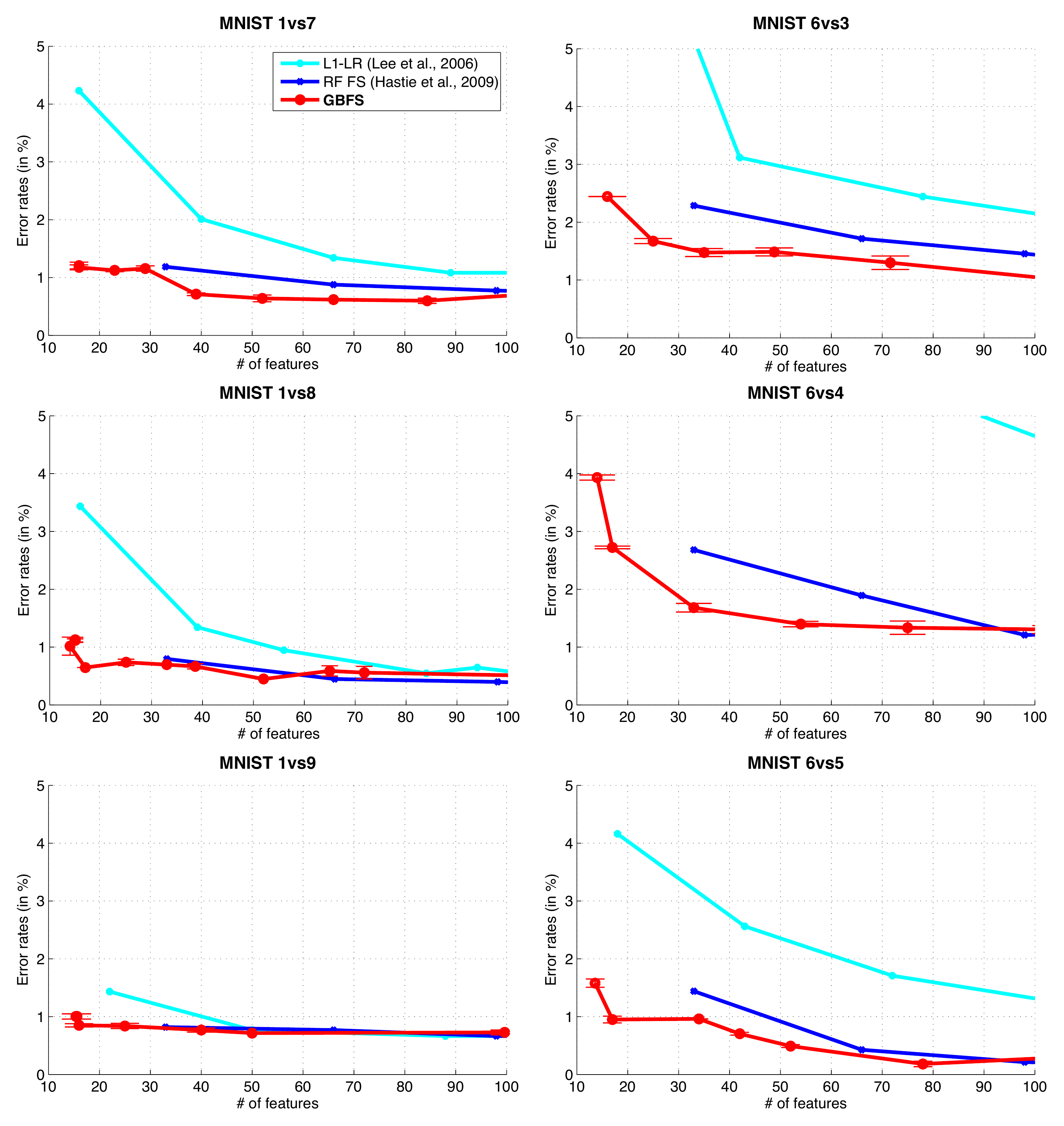}
}
\caption{Classification error rates vs.\ feature selection performance for different algorithms on 6 random chosen pairs of binary classification tasks from MNIST data set.}
\label{fig:mnistall}
\end{figure*}

Figure~\ref{fig:mnistall} indicates that \name{} consistently matches Random Forest FS, and clearly out-performs $l_1$-regularized logistic regression.  
\end{document}